\documentclass[12pt]{article}
\usepackage{geometry}

\usepackage{xcolor}         
\usepackage{latexsym}
\usepackage{amsmath}
\usepackage{amssymb}
\usepackage{times}
\usepackage{graphicx}
\usepackage{subfig}
\usepackage{epsfig}
\usepackage{array}
\usepackage{flafter}
\usepackage[page]{appendix}
\usepackage{listings}
\usepackage{algorithm,algpseudocode}
\usepackage{ulem}
\usepackage{multicol}
\usepackage{verbatim}
\usepackage{hyperref}
\usepackage{qtree}
\usepackage{tikz}
\usepackage{pgfplots}

\graphicspath{ {figures/} }

\newtheorem{definition}{Definition}
\newtheorem{theorem}{Theorem}
\newtheorem{ExampleDef}{Example}[section]

\newcommand{\Example}[3]{
  \begin{list}{}{
      \setlength{\leftmargin}{1em}} 
    \item                           
    \small                          
    \begin{ExampleDef} \rm          
      {\bf \hspace{-1ex}: #1}       
      #2                                
      \hfill {\large \boldmath $\Box$}  
      \label{ex:#3}                      
    \end{ExampleDef}
  \end{list}}

\pgfplotsset{compat=1.15}

\begin{document}
 
\begin{center}
{\Large \bf MinMax Networks \par} \vspace{1.5em}

{\large Winfried Lohmiller, Philipp Gassert, and Jean-Jacques Slotine \par}
{Nonlinear Systems Laboratory \\
Massachusetts Institute of Technology \\
Cambridge, Massachusetts, 02139, USA\\
{\sl wslohmil@mit.edu} \par}
\vspace{3em}
\end{center}

\begin{abstract}

While much progress has been achieved over the last decades in neuro-inspired machine learning, there are still fundamental theoretical problems in gradient-based learning using combinations of neurons. These problems, such as saddle points and suboptimal plateaus of the cost function, can lead in theory and practice to failures of learning. In addition, the discrete step size selection of the gradient is problematic since too large steps can lead to instability and too small steps slow down the learning. 

This paper describes an alternative discrete MinMax learning approach for continuous piece-wise linear functions. Global exponential convergence of the algorithm is established using Contraction Theory with Inequality Constraints \cite{Lohm2}, which is extended from the continuous to the discrete case in this paper: 
\begin{itemize}
    \item The parametrization of each linear function piece is, in contrast to deep learning, linear in the proposed MinMax network. This allows a linear regression stability proof as long as measurements do not transit from one linear region to its neighbouring linear region.
    \item The step size of the discrete gradient descent is  Lagrangian limited orthogonal to the edge of two neighbouring linear functions. It will be shown that this Lagrangian step limitation does not decrease the convergence of the unconstrained system dynamics in contrast to a step size limitation in the direction of the gradient. 
\end{itemize}
We show that the convergence rate of a constrained piece-wise linear function learning is equivalent to the exponential convergence rates of the individual local linear regions.
\end{abstract}

\section{Introduction} 

In this paper, we revisit standard convergence difficulties of gradient descent on a quadratic error cost, such as the possible presence of saddle points, sub-optimal plateaus, non-Lipschitz edges, time-varying measurements and the time discretization of the gradient.
Initial results of the MinMax learning we discuss were derived in \cite{Lohm4}\cite{Lohm5}.

The classical Rectified Linear Unit (ReLU) approach as e.g. in \cite{Hinton} implies a piece-wise linear approximation function.
The edges of the linear surfaces are not Lipschitz continuous. Hence it is difficult to prove stability and even uniqueness of the solution. 

The stability problem of the non-Lipschitz edges is overcome in this paper with a Lagrangian constraint step (\ref{eq:gj0dis}) to the edge. Since the edge belongs to both Lipschitz continuous regions it is possible at the next iteration to transit to the neighbouring region with a Lipschitz continuous step. Hence as a first step Contraction Theory of Constrained Continuous Systems \cite{Lohm2} is generalized in section \ref{discrete2} to discrete dynamics of the form \begin{equation}
  {\bf x}^{i+1} = {\bf f}({\bf x}^i, i)  \label{eq:f0dis}
\end{equation}
with $n$-dimensional discrete state $x^i$, time index $i$ and $l=1,..., L$-dimensional linear inequality constraints 
\begin{equation} 
g_l = {\bf g}_l^T (i+1){\bf x}^{i+1} + h_l(i+1)  \le  0. \label{eq:gj0dis}
\end{equation}
Note that many non-linear constraints can be brought into a linear form with a coordinate transformation, such that the results of this paper apply again.

Another problem in the stability analysis of deep learning is that e.g. a network of depth $100$ actually multiplies $100$ linear parameters with each other. Hence the smooth surface pieces are piece-wise polynomials of order $100$ w.r.t. the chosen parametrization, although the approximation function is actually piece-wise linear. To overcome this problem this paper suggests to use the sum of several piece-wise linear functions
\begin{equation}
\hat{y}({\bf x}) = \sum_{j=1}^{J_{\min}} \hat{y}_{j \min}({\bf x}) + \sum_{j=J_{\min}}^{J_{\max}} \hat{y}_{j \max}({\bf x}) 
\label{eq:minmax}
\end{equation}
which uses the convex and concave neurons $j$
\begin{eqnarray}
\hat{y}_{j \min}({\bf x}) &=& \min(\hat{z}_{j1}, ..., \hat{z}_{j K_j}) \nonumber  \\
\hat{y}_{j \max}({\bf x}) &=& \max(\hat{z}_{j1}, ..., \hat{z}_{j K_j}) \nonumber
\end{eqnarray}
which consist of $k = 1, ..., K_j$ linear basic neurons $\hat{z}_{jk} = {\bf x}^T \hat{\bf w}_{jk}$ of estimated $N+1$-dimensional weight vector $\hat{\bf w}_{jk}$.

The concave $\max$ and convex $\min$ functions are a direct generalization of the ReLU to achieve piece-wise linear functions. Multiple local convex and concave functions can be approximated with multiple $\min$ and $\max$ operators. The key advantage to deep networks is that the parametrization is still linear in $\hat{\bf w}_{jk}$ between the edges, i.e. linear stability proofs can be used with the mentioned step size limitation to the edge.

The following example shows the main modeling difference of both approaches:

\Example{}{Let us consider the unit pyramid in subfigure \ref{subfig:a} 
\begin{eqnarray}
y({\bf x}) &=& \max( 0, \min(x_1 + 1, x_2 + 1, -x_1+1, -x_2+1) ). \nonumber 
\end{eqnarray}
In a deep ReLU network each ReLU adds an edge to the piece-wise linear function.
A ReLU network estimation of depth 2 exactly models the pyramid above with
\begin{eqnarray}
\hat{y}({\bf x}) &=& \max(0, \hat{z}_{11} ) \nonumber \\
\hat{z}_{11} &=& 1 - 0.5 \left( \hat{y}_{21} + \hat{y}_{22} + \hat{y}_{23} +\hat{y}_{24} \right) ) \nonumber \\
\hat{y}_{2j} &=& \max(0, \hat{z}_{2j}) \nonumber \\
\hat{z}_{21} &=& -x_1-x_2, \
\hat{z}_{22} = x_1-x_2, \
\hat{z}_{23} = x_1+x_2, \
\hat{z}_{24} = -x_1+x_2 \nonumber
\end{eqnarray}
The lower ReLUs $\hat{y}_{2j} = \max(0, \hat{z}_{2j})$ define the the 4 edges of the pyramid without ground leading to the linear input $\hat{z}_{11}$ to the last layer in subfigure \ref{subfig:b}. The upper ReLU $\hat{y} = \max(0, \hat{z}_{11} )$ then adds the remaining 4 edges to the ground. In general a high over-parametrization is needed to approximate a piece-wise linear functions with deep ReLUs \cite{Hinton}. Also it is not very easy to understand which edge is defined by which ReLU in which layer.

In contrast to the above the proposed MinMax approach (\ref{eq:minmax}) systematically defines 
\begin{itemize}
    \item all convex edges of the pyramid in $\hat{y}_{1 \max}$ and 
    \item all concave edges of the pyramid in $\hat{y}_{2 \min}$ 
\end{itemize}
with
\begin{eqnarray}
\hat{y}({\bf x}) &=& \hat{y}_{1 \max} + \hat{y}_{2 \min} \nonumber \\ 
\hat{y}_{1 \max} &=& \max(\hat{z}_{11}, \hat{z}_{12}, \hat{z}_{13}, \hat{z}_{14}, \hat{z}_{15}), \ \hat{y}_{2 \min} = \min(\hat{z}_{21}, \hat{z}_{22}, \hat{z}_{23}, \hat{z}_{24}) \nonumber \\ 
\hat{z}_{11} &=& 0, \
\hat{z}_{12} = x_1 + 1, \
\hat{z}_{13} = x_2 + 1, \
\hat{z}_{14} = -x_1+1, \
\hat{z}_{15} = -x_2+1 \nonumber  \\
\hat{z}_{21} &=& x_1 + 1, \
\hat{z}_{22} = x_2 + 1, \
\hat{z}_{23} = -x_1+1, \ 
\hat{z}_{24} = -x_2+1  \nonumber
\end{eqnarray}
where subfigures \ref{subfig:c} and \ref{subfig:d} show the convex and concave neurons of the MinMax approach. The legend indicates which basic neurons produce a nonzero output for the colored surfaces. 

\begin{figure*}
  \subfloat[Unit pyramid]{\includegraphics[width=0.49\textwidth]{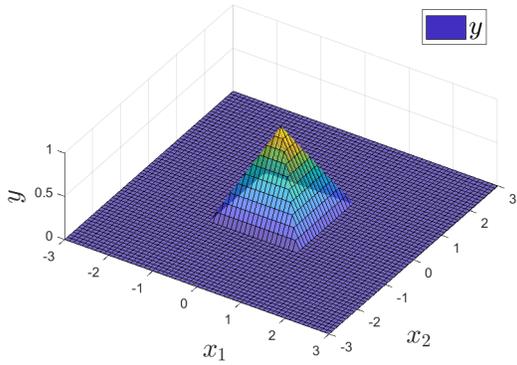} \label{subfig:a}}
  \subfloat[Intermediate neuron $\hat{z}_1$ of ReLU network]{\includegraphics[width=0.49\textwidth]{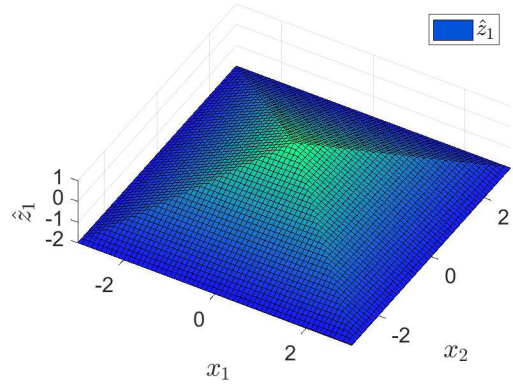} \label{subfig:b}} \par
  \subfloat[Concave neuron $\hat{y}_{1 \max}$ of MinMax]{\includegraphics[width=0.49\textwidth]{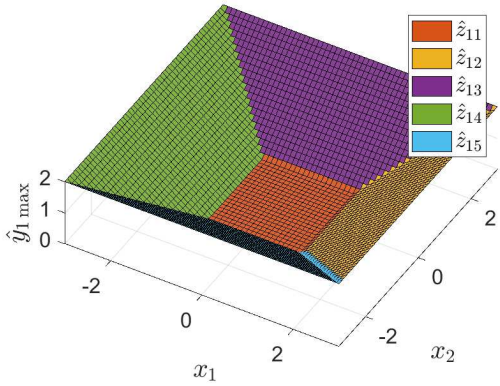} \label{subfig:c}}
  \subfloat[Convex neuron $\hat{y}_{2 \min}$ of MinMax]{\includegraphics[width=0.49\textwidth]{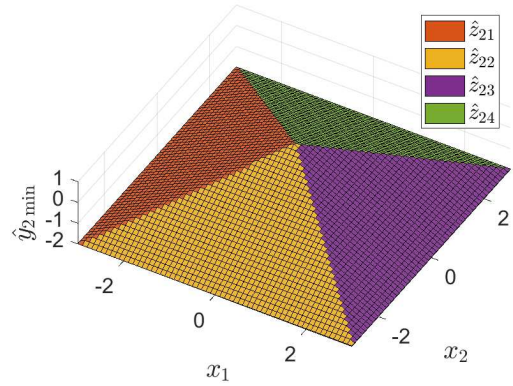} \label{subfig:d}}
  \caption{Unit pyramid, intermediate ReLU neuron and a MinMax network with two neurons} \label{fig:results}
\end{figure*}
}{20}

Motivated by the above section \ref{jndis} introduces a piece-wise linear MinMax discrete function learning (\ref{eq:minmax}) for the $N$-dimensional case. The approach still uses gradient descent on a quadratic cost in discrete time. Exact exponential stability guarantees are provided using the results of section \ref{discrete2}. Saddle points or sub-optimal plateaus are avoided with a linear parametrization. Possible instabilities of the non-Lipschitz edges are avoided with intermediate Lagrange constraints (\ref{eq:gj0dis}). Note that time-varying measurements and the time discretization of the gradient are part of the exponential stability proof. Finally the result is summarised in the Summary.

\section{Discrete-time constrained systems} \label{discrete2}

This section extends Contraction Analysis of Continuous Constrained Systems \cite{Lohm2} to the discrete case: 

A constraint $l$ in (\ref{eq:gj0dis}) only has an impact on the unconstrained dynamics (\ref{eq:f0dis}) if the inequality turns into an equality $g_l = 0$ , leading to the following definition.
\begin{definition} 
The set of active constraints $\mathcal{A}(({\bf x}^{i+1}, {i+1})  \subseteq \{1, ..., L \}$ contains the elements $l$ (\ref{eq:gj0dis}) which are on the boundary of the original constraint
\begin{eqnarray}
g_l = {\bf g}_l^T (i+1){\bf x}^{i+1} + h_l(i+1)  &=& 0. \nonumber
\label{def:equalconstraintdis}
\end{eqnarray}
\end{definition}
The constrained dynamic equations (\ref{eq:f0dis}, \ref{eq:gj0dis}) are then of the form
\begin{equation}
  {\bf x}^{i+1} = {\bf f}({\bf x}^i, i) + \sum_{all \ l \in \mathcal{A}} {\bf g}_l {\lambda}_l.   \label{eq:xdiff}
\end{equation}
All constraints $g_l, l \in \mathcal{A}$ are not violated at $i+1$ if
$$
 {g}_l = {\bf g}^T_l(i+1) \left( {\bf f}({\bf x}^i, i) + \sum_{all \ l \in \mathcal{A}} {\bf g}_l(i+1) {\lambda}_l \right) + h_l(i+1) \le  0
$$
leading to the following definition: 
\begin{definition} 
The set of Lagrange multipliers $\lambda_l, l \in \mathcal{A}$ of (\ref{eq:gj0dis}) is the solution of the linear programming problem 
\begin{eqnarray}
 maximize \sum_{all \ l \in \mathcal{A}} \lambda_l \ subject \ to &&  \nonumber \\
{\bf g}^T_l \left( {\bf f}({\bf x}^i, i) + \sum_{all \ l \in \mathcal{A}} {\bf g}_l {\lambda}_l \right) + h_l &\le& 0 \nonumber \\
\lambda_l &\le& 0 \nonumber 
\end{eqnarray}
 \label{def:activeconstraintdis}
\end{definition}
where we requested similar to  \cite{Lohm2} that the constraint term points in the interior of the constraint with $\lambda_l \le 0$ and the negative $\lambda_l$ are maximized \cite{Bryson} to minimize their joint usage. The equation above corresponds to the initial solution problem of Linear Programming (LP), see e.g. section LP in \cite{Bronstein}. Diverse solutions exist for this problem, such as the simplex algorithm in section LP in \cite{Bronstein}.

Similar to \cite{Lohm2} we introduce a virtual displacement between two neighbouring trajectories, which is constrained by $g_l = 0$ at the active $l \in \mathcal{A}$. This virtual displacement has to be parallel to $g_l = 0$, i.e. orthogonal to the normals ${\bf g}_l$, which implies 
\begin{eqnarray}
{\bf \delta x}^{i+1} &=& {\bf G}_{\parallel}(i+1) {\bf \delta x}^{i+1 \ast}  \label{eq:constvirtMdiff} \\
 {\bf G}_{\parallel}^T {\bf G}_{\parallel} &=& {\bf I}  \nonumber \\
{\bf G}_{\parallel}^{T} {\bf g}_l(i+1) &=& {\bf 0} \ \ \ \ \ \ \forall \ l \in \mathcal{A}  \nonumber
\end{eqnarray}
where ${\bf \delta x}^{i+1 \ast}$ is the reduced virtual displacement of dimension $n$ minus by the number of active constraints in $\mathcal{A}$. 

The constrained dynamic equations (\ref{eq:xdiff}) can be rewritten as:
\begin{equation}
  {\bf x}^{i+1} = {\bf f}({\bf x}^i, i) + \sum_{all \ l \in \mathcal{A}} {\bf g}_l \lambda_l = {\bf f}({\bf x}^i, i) + \sum_{l=1}^L  {\bf g}_l step(g_l) \lambda_l \nonumber
\end{equation}
with the step function
\begin{equation}
step(g_l) =
\left\{ 
\begin{array}{l}
0 \text{\ \ for \ \ } g_l < 0  \\
1 \text{\ \ for \ \ } g_l = 0 \\
undefined \text{\ \ for \ \ } g_l > 0 
\end{array}
\right. \nonumber
\end{equation}
whose variation is
\begin{eqnarray}
\delta  {\bf x}^{i+1} &=& \frac {\partial {\bf f}} {\partial {\bf x}^i} {\bf \delta x}^i + \sum_{l = 1}^L \left( {\bf g}_l \frac{\partial \lambda_l}{\partial {\bf x}^{i+1}} + {\bf g}_l \frac{\partial step(g_l)}{\partial g_l} {\bf g}_l^T \lambda_l \right)  {\bf \delta x}^{i+1} \nonumber 
\end{eqnarray}
Now the squared virtual length dynamics with a metric ${\bf M}^i( {\bf x}^i, i)$ can be bounded as
\begin{eqnarray}
\frac{1}{2}\  {\bf \delta x}^{i+1 T} {\bf M}^{i+1} {\bf \delta x}^{i+1} &=& 
{\bf \delta x}^{i T} \frac {\partial {\bf f}} {\partial {\bf x}^i}^T {\bf M}^{i+1} \frac {\partial {\bf f}} {\partial {\bf x}^i}  {\bf \delta x}^i + \frac{1}{2}\  {\bf \delta x}^{i+1 T}  ({\bf M}^{i+1} {\bf g}_l \frac{\partial step(g_l)}{\partial g_l} {\bf g}_l^T )_H\lambda_l {\bf \delta x}^{i+1} \nonumber 
\end{eqnarray}
where we used ${\bf G}_{\parallel}^T {\bf g}_l \frac{\partial \lambda_l }{\partial {\bf x}^{i+1}} = {\bf 0}$ since on the constraint the first two terms vanish and outside the constraint the last term vanishes. The Dirac impulse $\frac{\partial step(g_l)}{\partial g_l}$ discontinuously sets the virtual displacement ${\bf \delta x}^{i+1}$ to ${\bf G}_{\parallel} {\bf \delta x}^{i+1 \ast}$  when a constraint is activated.

Thus the dynamics of ${\bf \delta x}^{i+1 T}  {\bf \delta x}^{i+1} $ is composed of exponentially convergent continuous segments and an enforcement of ${\bf \delta x}^{i+1}$ to ${\bf G}_{\parallel} {\bf \delta x}^{i+1 \ast}$ at the activation of a constraint.
Let us summarize this result.
\begin{theorem}
Consider the discrete dynamics
\begin{equation}
  {\bf x}^{i+1} = {\bf f}({\bf x}^i, i) + \sum_{all \ l \in \mathcal{A}} {\bf g}_l \lambda_l  \label{eq:xdiff2}
\end{equation}
within the metric ${\bf M}({\bf x}^i, i)$ constrained by a $l=1, ..., L$-dimensional inequality constraint
\begin{equation}
 g_l = {\bf g}_l^T (i+1){\bf x}^{i+1} + h_l(i+1)  \le 0 \label{eq:gjconsdi}
\end{equation}
The set of active constraints $\mathcal{A}$ and Lagrange multipliers $\lambda_{l}$ are given in Definition \ref{def:equalconstraintdis} and \ref{def:activeconstraintdis}

The distance $s = \min \int_{{\bf x}^i(s) = {\bf x}^i_1}^{{\bf x}^i_2} \sqrt{ \delta {\bf x}^{i T}  {\bf M}^i \delta {\bf x}^i }$ within $\mathbb{G}^n$ from any trajectory ${\bf x}^i_1(t)$ to any other trajectory ${\bf x}^i_2(t)$ converges exponentially to $0$ with an exponential convergence rate $\le \max_{along \ s} (\sigma_{\max}({\bf x}^i, i)), \sigma_{\max} > 0$ ($\ge \min_{along \ s} (\sigma_{\min}({\bf x}^i, i)), \sigma_{\min} > 0$) with
\begin{eqnarray}
\sigma_{\min}^2  {\bf G}^{T}_{\parallel} {\bf M}^{i}  {\bf G}^{T}_{\parallel} \le {\bf G}_{\parallel} \frac {\partial {\bf f}} {\partial {\bf x}^i}^T  {\bf M}^{i+1}  \frac {\partial {\bf f}} {\partial {\bf x}^i} {\bf G}^{T}_{\parallel}  &\le& \sigma_{\max}^2  {\bf G}^{T}_{\parallel} {\bf M}^{i}  {\bf G}^{T}_{\parallel}  \label{eg:genJacdis}
\end{eqnarray} 
 with the constrained tangential space ${\bf G}_{\parallel}$ from equation (\ref{eq:constvirtMdiff}).
 \label{th:theoremFdis}
 
In addition the activation of a constraint discontinuously sets the virtual displacement ${\bf \delta x}^{i+1}$ to ${\bf G}_{\parallel} {\bf \delta x}^{i+1 \ast}$.
\end{theorem}
Note that section 3.3. of \cite{Bertesekas} provides a stability condition on the constrained step ${\bf f}_{i+1} - {\bf f}_i$ for the original contraction mapping theorem. Theorem \ref{th:theoremFdis} makes this condition more concrete by giving explicit conditions on the derivatives of ${\bf f}$. It also extends the result to a metric.
In addition all notes of the continuous theorem in \cite{Lohm2} apply here as well.

\section{Discrete exponential stable learning}\label{jndis}

Let us us assume a piece-wise linear $N$-dimensional measurement function
\begin{equation}
y = y({\bf x}) \nonumber
\end{equation}
where the $N$-dimensional input vector ${\bf x}'$ is augmented to the $N+1$-dimensional input vector ${\bf x} = ({\bf x}^{'T}, 1)^T$. We have $m=1,...,M$ measurements 
\begin{equation}
y_m^i = y({\bf x}_m^i) \nonumber
\end{equation}
with the measured input vector ${\bf x}^i_m$ at time index $i$. The goal is to approximate $y_m^i$ with $\hat{y}_m^{i}({\bf x}_m^{i})$. We achieve the approximation of the true function by minimizing the weighted cost
\begin{equation}
V = \frac{1}{2} \sum_{m =1}^M \alpha_m^2 \tilde{y}_m^{i 2}  \nonumber
\end{equation}
with $\tilde{y}_m^i  = \hat{y}_m^i({\bf x}_m^i) - y^i_m({\bf x}_m^i)$ and measurement weight $\alpha_m^2({\bf x}_m^i) \ge 0$. The unconstrained parameter learning law is the classical gradient descent of $V$
\begin{eqnarray}
{\hat{\bf W}}_{j}^{i+1} &=& {\hat{\bf W}}_{j}^{i} -  \frac{\partial V}{\partial \hat{\bf W}_{j}} \nonumber \\
&=& {\hat{\bf W}}_{j}^{i} -   \sum_{m =1}^M   {\bf A}_{j}({\bf x}_m^i) \alpha_m^2 \tilde{y}_m \nonumber \\
&=& {\hat{\bf W}}_{j}^{i} -   \sum_{m =1}^M  {\bf A}_{j}({\bf x}_m^i) \alpha_m^2 \left(\sum_{l=1}^{J} {\bf A}_{l}^T({\bf x}_m^i)  {\hat{\bf W}}_{l}^{i} -y^i_m \right) \label{eq:unconlear} 
\end{eqnarray}
with $\hat{\bf W}_j^{i T} = (\hat{\bf w}_{j1}^{i T}, ..., \hat{\bf w}_{jK_j}^{i T})$ and the activation function
\begin{equation}
{\bf A}_{j} ({\bf x})= 
\left\{ 
\begin{array}{l}
{\bf x} \text{\ \ if } \hat{z}_{jk{\ast}}= \hat{y}_{j \min} ( \hat{y}_{j \max} ) \text{ for a } \min (\max) \text{ neuron} \\
{\bf 0} \text{\ \ for all other \ } k \ne  k^{\ast}
\end{array}
\right. \label{eq:activation}
\end{equation}
where for multiple solutions $\hat{z}_{jk{\ast}}= \hat{y}_{j \min} ( \hat{y}_{j \max} )$ only one activation is set to ${\bf x} $ and all others to ${\bf 0}$. Note that the gradient in (\ref{eq:unconlear}) could be multiplied on top with a gain $\alpha_j(i)$ which we have skipped for simplicity here.

The weight dynamics (\ref{eq:unconlear}) is equivalent to the measurement estimation dynamics
\begin{eqnarray}
\hat{y}^{i+1}_n &=& \sum_{j =1}^{J}  {\bf A}_{j}^T({\bf x}_n^{i+1}) {\hat{\bf W}}_{j}^{i+1} \nonumber \\
&=& \hat{y}^{i}_n- \sum_{j =1}^{J} {\bf A}_{j}^T({\bf x}_n^i)   \sum_{m =1}^M   {\bf A}_{j}({\bf x}_m^i) \alpha_m^2 \tilde{y}_m \label{eq:yi1} .
\end{eqnarray}
In this last formulation we assume static measurements ${\bf x}_n^{i+1} = {\bf x}_n^{i}$.

To assure Lipschitz continuity of the active basic neurons $\hat{z}_{jk^{\ast}}$ at the learning step $i+1$ we have to exclude with Theorem \ref{th:theoremFdis} a transition beyond the edges of the $\min$ and $\max$ operator with the constraints
\begin{eqnarray}
    g_{l} = {\bf g}_{l}^{T} \hat{\bf W}_{j}^{i+1} = {\bf x}_m^{i+1 T} \hat{\bf w}_{j k^{\ast}}^{i+1} - {\bf x}_m^{i+1 T} \hat{\bf  w}_{jk}^{i+1} &\le& 0 \text{ for a } \min \text{ neuron j and all } k \ne k^{\ast} \nonumber \\
    g_{l} = {\bf g}_{l}^{T} \hat{\bf W}_{j}^{i+1} = {\bf x}_m^{i+1 T} \hat{\bf w}_{jk}^{i+1} - {\bf x}_m^{i+1 T} \hat{\bf w}_{j k^{\ast}}^{i+1} &\le& 0 
  \text{ for a } \max \text{ neuron j and all } k \ne k^{\ast} \nonumber
\end{eqnarray}
where the related Lagrange parameters are given in Definition \ref{def:activeconstraintdis}. 

Theorem \ref{th:theoremFdis} then implies global contraction behaviour with the largest and smallest singular value of the variation of (\ref{eq:unconlear}) or alternatively (\ref{eq:yi1}) within the constraints above.

Summarizing the above leads to:
\begin{theorem}
\itshape
Consider a piece-wise linear $N$-dimensional measurement function
\begin{equation}
y = y({\bf x}) \label{eq:yx}
\end{equation}
where the $N$-dimensional input vector ${\bf x}'$ is augmented to the $N+1$-dimensional input vector ${\bf x} = (1, {\bf x}^{'T})^T$. We have $m=1,...,M$ measurements 
\begin{equation}
y_m^{i} = y({\bf x}_m^{i}) \label{eq:yxk}
\end{equation}
with the measured input vector ${\bf x}_m^{i}$ at time index $i$. We approximate (\ref{eq:yx}) with 
\begin{equation}
\hat{y}({\bf x}_m^{i}) = \sum_{j=1}^{J_{\min}} \hat{y}_{j \min}({\bf x}_m^{i}) + \sum_{j=J_{\min}}^{J_{\max}} \hat{y}_{j \max}({\bf x}_m^{i}) \label{eq:hatyx}
\end{equation}
which uses the convex and concave neurons $j$
\begin{eqnarray}
\hat{y}_{j \min}({\bf x}_m^{i}) &=& \min(\hat{z}_{j1}, ..., \hat{z}_{j K_j}) \nonumber  \\
\hat{y}_{j \max}({\bf x}_m^{i}) &=& \max(\hat{z}_{j1}, ..., \hat{z}_{j K_j}) \label{eq:neuronj}
\end{eqnarray}
which consist of $k = 1, ..., K_j$ linear basic neurons $\hat{z}_{jk} = {\bf x}_m^{i T} \hat{\bf w}_{jk}^i$ of estimated $N+1$-dimensional weight vector $\hat{\bf w}_{jk}^i$ at time $i$ and the activation function
(\ref{eq:activation}). For a neuron $j$ we constrain the dynamics with 
\begin{eqnarray}
  g_{l} = {\bf g}_{l}^{T} \hat{\bf W}_{j}^{i+1} = {\bf x}_m^{i+1 T} \hat{\bf w}_{j k^{\ast}}^{i+1} - {\bf x}_m^{i+1 T} \hat{\bf  w}_{jk}^{i+1} &\le& 0 \text{ for a } \min \text{ neuron j, all } k \ne k^{\ast} \nonumber \\
   g_{l} = {\bf g}_{l}^{T} \hat{\bf W}_{j}^{i+1} = {\bf x}_m^{i+1 T} \hat{\bf w}_{jk}^{i+1} - {\bf x}_m^{i+1 T} \hat{\bf w}_{j k^{\ast}}^{i+1} &\le& 0 
  \text{ for a } \max \text{ neuron j, all } k \ne k^{\ast}
  \label{eq:xkxkdisx}
\end{eqnarray}
with $\hat{\bf W}_j^{iT} = (\hat{\bf w}_{j1}^{i T}, ..., \hat{\bf w}_{jK_j}^{iT})$. The constrained learning of the cost 
\begin{equation}
  V = \frac{1}{2} \sum_{m =1}^M \alpha_m^2 \tilde{y}_m^{i 2} \label{eq:V}  
\end{equation}
with $\tilde{y}_m^i  = \hat{y}_m^i({\bf x}_m^i) - y^i_m({\bf x}_m^i)$ and measurement weight $\alpha_m^2({\bf x}_m^i) \ge 0$ 
\begin{eqnarray}
\hat{\bf W}_{j}^{i+1} &=& {\hat{\bf W}}_{j}^{i} -  \frac{\partial V}{\partial \hat{\bf W}_{j}} + \sum_{all \ l \in \mathcal{A}}  {\bf g}_{l} \lambda_l \nonumber \\
&=&   \hat{\bf W}_{j}^{i} - \sum_{m =1}^M  {\bf A}_{j}({\bf x}_m)  \alpha_m^2(i) \tilde{y}_m^i + \sum_{all \ l \in \mathcal{A}}  {\bf g}_{l} \lambda_l 
\label{eq:learning1Dv2}
\end{eqnarray}
where the set of active constraints $\mathcal{A}$ and Lagrange multipliers $\lambda_l$ are given in Definition \ref{def:equalconstraintdis} and \ref{def:activeconstraintdis} is globally exponentially converging to $ \frac{\partial V}{\partial \hat{\bf W}_{j}} = {\bf 0}$ with the largest and smallest singular value of the matrix 
\begin{eqnarray}
{\bf I}_{jk}  -  \sum_{m =1}^M  {\bf A}_{j}({\bf x}_m^i) \alpha_m^2 {\bf A}_{k}^T({\bf x}_m^i) 
\label{eq:Oii}
\end{eqnarray} 
or alternatively in the metric $\alpha_n^2$ with the block diagonal matrix
\begin{eqnarray}
{\bf I}_{nm}  -  \sum_{j =1}^{J} \alpha_n {\bf A}_{j}^T({\bf x}_n^i) {\bf A}_{j}({\bf x}_m^i) \alpha_m \label{eq:ii1al} 
\end{eqnarray}
In this last formulation we assume static measurements ${\bf x}_n^{i+1} = {\bf x}_n^{i}$  e.g. for batch processing. 

The equilibrium point $ \frac{\partial V}{\partial \hat{\bf W}_{j}} = {\bf 0}$ is unique or global if the largest singular value is strictly less then $1$.
\label{th:1Dv2dis}
\end{theorem}
Note that e.g. $0 < \alpha_m \le \frac{1}{J \lvert {\bf x}_m^{i} \rvert}$ assures contraction behaviour in (\ref{eq:ii1al}).

Also note that at the minimum of $V$ the cost $V$ will not be $0$ if an incomplete active topology (\ref{eq:hatyx}) was used. Hence  the following pruning and creation principles have to be applied to find a right active topology (\ref{eq:hatyx}). For a right topology $V$ will go to $0$ exponentially since a contracting solution $V=0$ then exists.
\begin{itemize}
\item Inactive neurons (\ref{eq:neuronj}) close to $0$ can be pruned.

\item Basic neurons of a neuron, which never become active or which are similar to another basic neuron, can be pruned.

\item New convex or concave neurons (\ref{eq:neuronj}) $j$ can be initialized with all $\hat{z}_{jk} = 0$ and one activated basic neuron if persistent relevant errors $\tilde{y}_m$ remain. 

In principle one convex and one concave neuron are sufficient to activate convex or concave edges everywhere. However, the numerical complexity of the simplex algorithm in Definition \ref{def:activeconstraintdis} is exponential in the worst case. Hence it is computationally more efficient to take several convex or concave neurons of low dimension rather then one convex or concave neuron of very high dimension.

\item Similarly, new basic neurons can be created by duplicating existing basic neurons $jk$ with persistent relevant errors $\tilde{y}_m$ 
$$
\hat{z}_{j new} = \hat{z}_{jk} 
$$ 
\end{itemize}
Note that when the Lagrangian constraint is active at the boundary between two basic neurons the following cases exist for the trajectory which starts at the next iteration on the boundary between both neurons:
\begin{itemize}
 \item Both trajectories move away from the boundary. Here one trajectory can be selected. Several solutions may exist here since the boundary is not Lipschitz continuous. Note that especially in this case $\delta  {\bf x}^{i+1}$ does not diverge since it was parallelized to the constraint the iteration before.
\item One trajectory moves to the boundary the other away from the boundary. Here the trajectory which moves away from the boundary is selected.
\item Both trajectories again move to the boundary. Here the Lagrangian constraint has to be maintained active at the next iteration, such that the trajectory moves on the edge until it eventually leaves the edge at a later iteration.
\end{itemize}
The following example illustrates the effect above:
\Example{}{Figure \ref{fig:learningex} shows the learning of a single measurement at the edge of 2 basic neurons of Theorem  \ref{th:1Dv2dis}:
 \begin{itemize}
 \item The left side shows a measurement on the convex side of 2 basic neurons. Learning of the left or right neuron alone does not work since the measurement cost (\ref{eq:V}) at $i$ 
\begin{equation}
V = \frac{1}{2} \sum_{m =1}^M \tilde{y}_m^{i 2} \nonumber
\end{equation}
has different measurements $m$ at $i+1$. Learning works here only with an active Lagrangian constraint (\ref{eq:xkxkdisx}) of Theorem \ref{th:1Dv2dis}

\item On the right side the measurement is on the concave side of 2 basic neurons. Here both neurons can learn without an active Lagrangian constraint (\ref{eq:xkxkdisx}) since the measurement stays at $i+1$ at the same basic neuron where it was at $i$.
 \end{itemize}
\begin{figure*}
\begin{center}
\includegraphics[scale=0.52]{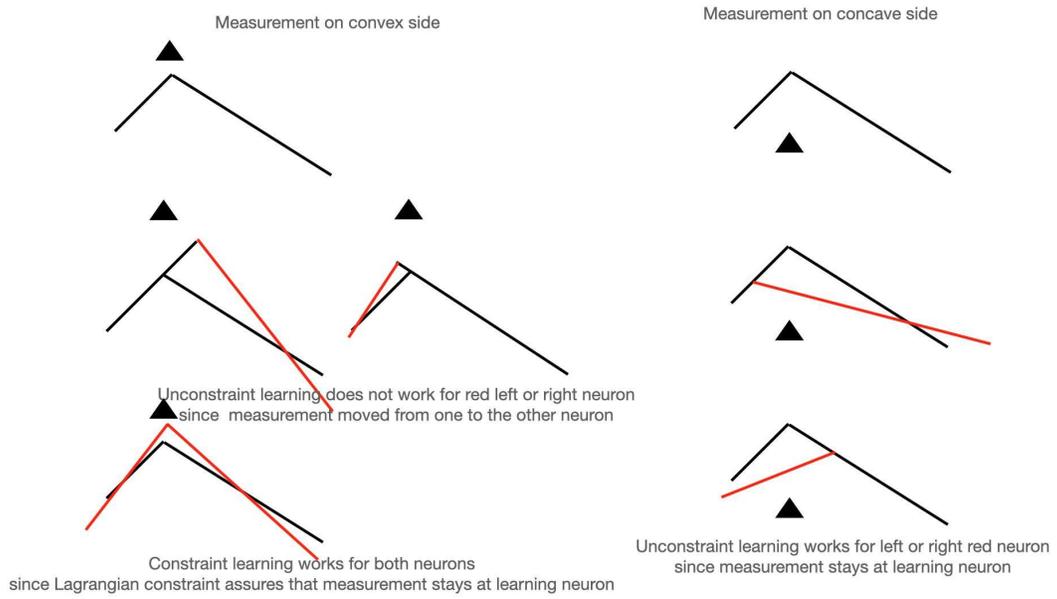}
\end{center}
\caption{Learning with measurements on convex or concave side}
\label{fig:learningex}
\end{figure*}
}{xxx}

\Example{}{Figure \ref{fig:ply} shows how a MinMax network of Theorem \ref{th:1Dv2dis} evolves for a polygon. We approximate the target polygon $y$ (\ref{eq:yx}) in subfigure \ref{subfig:a_1D} with the MinMax learning (\ref{eq:learning1Dv2}):
\begin{itemize}
    \item We start with one linear neuron $\hat{y} = y_{1,min} = min(\hat{z}_{11})$ \eqref{eq:neuronj}  in subfigure \ref{subfig:b_1D}. 
    \item We then insert a new basic neuron in $\hat{y} = y_{1,min} = min(\hat{z}_{11}, \hat{z}_{12})$, where $\hat{z}_{11}$ and $\hat{z}_{12}$ have initially the same parameters. After training, the network converges to the approximation in subplot \ref{subfig:c_1D}.
    \item A new concave neuron with two basic neurons is then inserted leading to $\hat{y} = min(\hat{z}_{11},\hat{z}_{12}) + max(\hat{z}_{21}, \hat{z}_{22})$ in subfigure \ref{subfig:d_1D}. 
    \item After another insertion of a basic neuron, we get a perfect approximation in subfigure \ref{subfig:e_1D} with $\hat{y} = min(\hat{z}_{11},\hat{z}_{12}) + max(\hat{z}_{21}, \hat{z}_{22}, \hat{z}_{23})$. The final two neurons (\ref{eq:neuronj}) are depicted in subfigures \ref{subfig:f_1D} and \ref{subfig:g_1D}.
\end{itemize}
The above example shows how important the (basic) neuron creation is to learn the topology of the MinMax network.

The approach of this paper uses 5 basic neurons, whereas the benchmark in \cite{Shai} had 100 to 2000 neurons in several layers. The benchmark in \cite{Shai} took up to 50000 iterations to converge to remaining persistent errors, where the MinMax network only needs a few hundred. The approach of this paper needs 8 measurement points, i.e. the minimum number of points to define the polygon. \cite{Shai} used 100 measurement points for learning.

\begin{figure*}
\thispagestyle{empty} 
\centering
  \subfloat[Target function $y$]{\centering \includegraphics[width=0.35\textwidth]{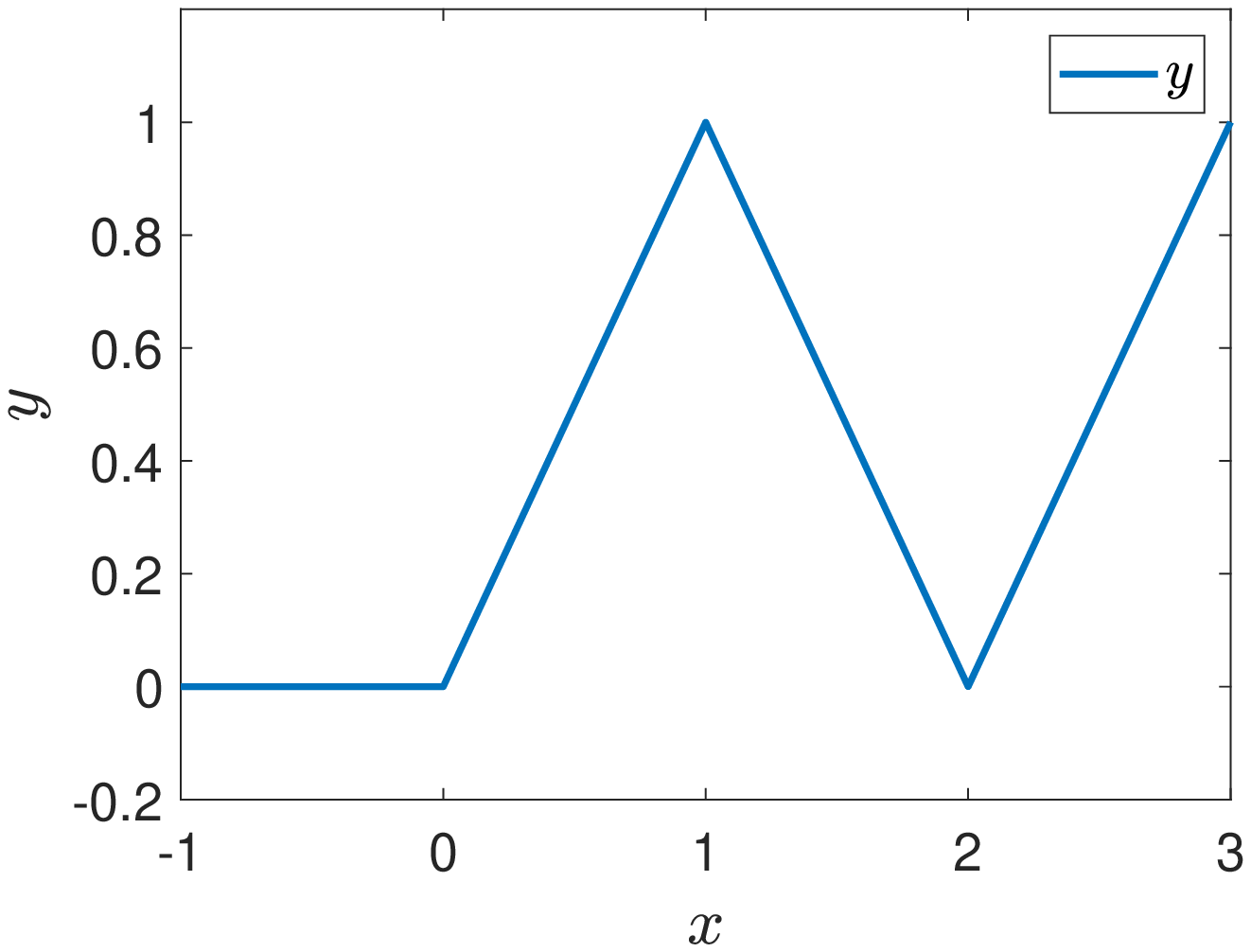} \label{subfig:a_1D}} \par
  \vspace{-10pt}
  \subfloat[$\hat{y}$ with one linear neuron]{\includegraphics[width=0.35\textwidth]{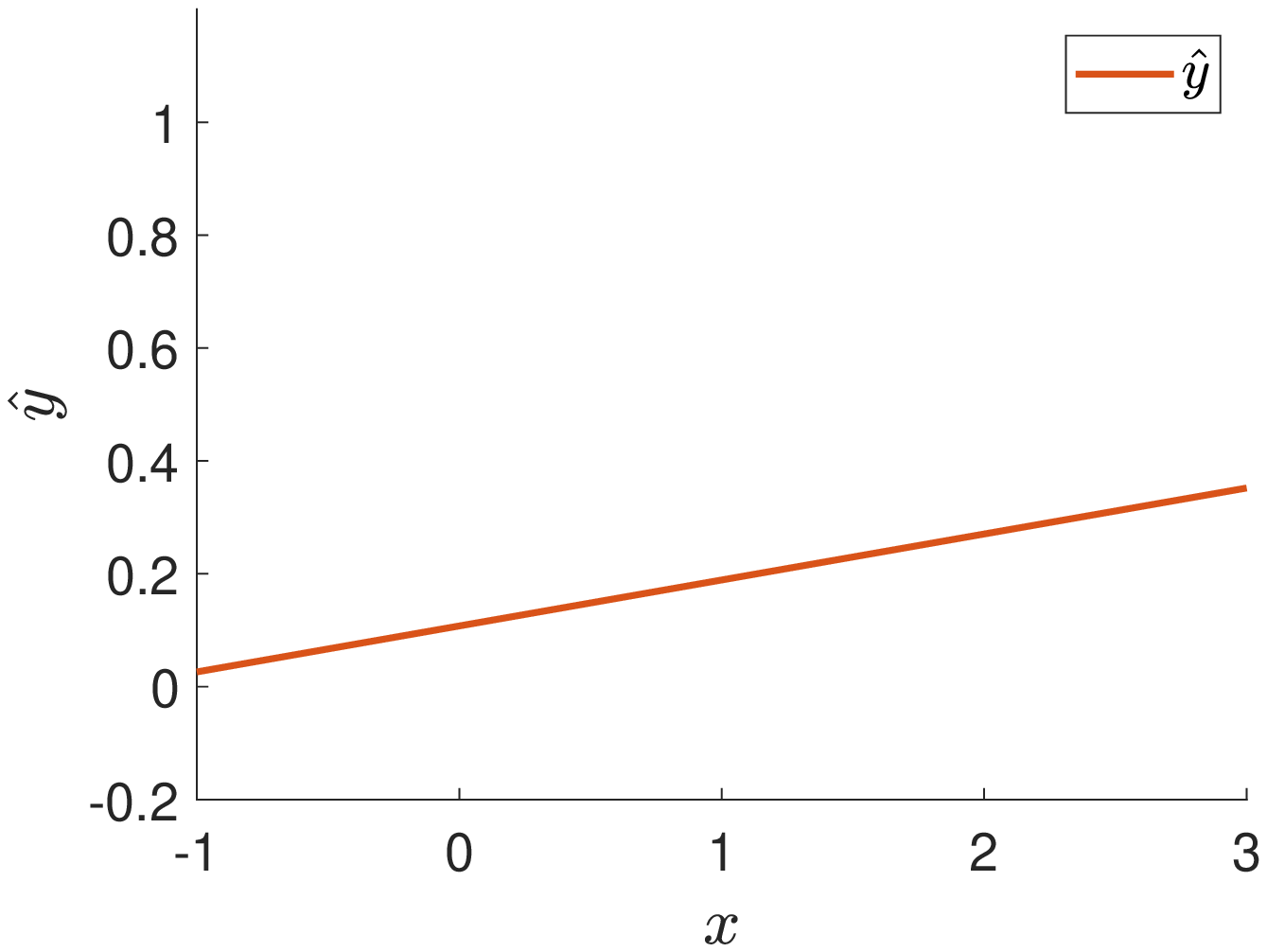} \label{subfig:b_1D}}
  \subfloat[$\hat{y}$ with one convex neuron]{\includegraphics[width=0.35\textwidth]{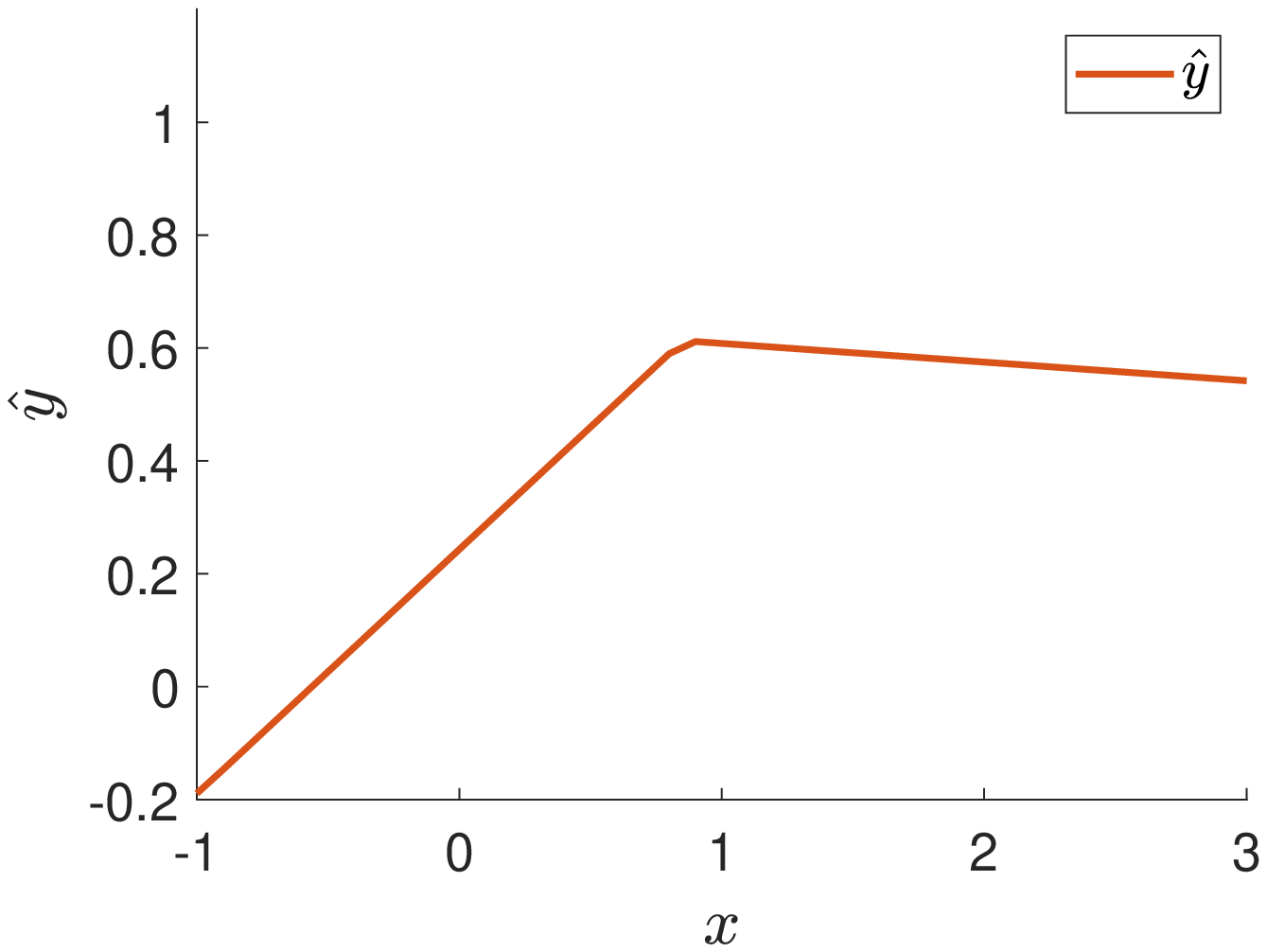} \label{subfig:c_1D}} \par
  \vspace{-10pt}
  \subfloat[$\hat{y}$ with one concave and one convex neuron ]{\includegraphics[width=0.35\textwidth]{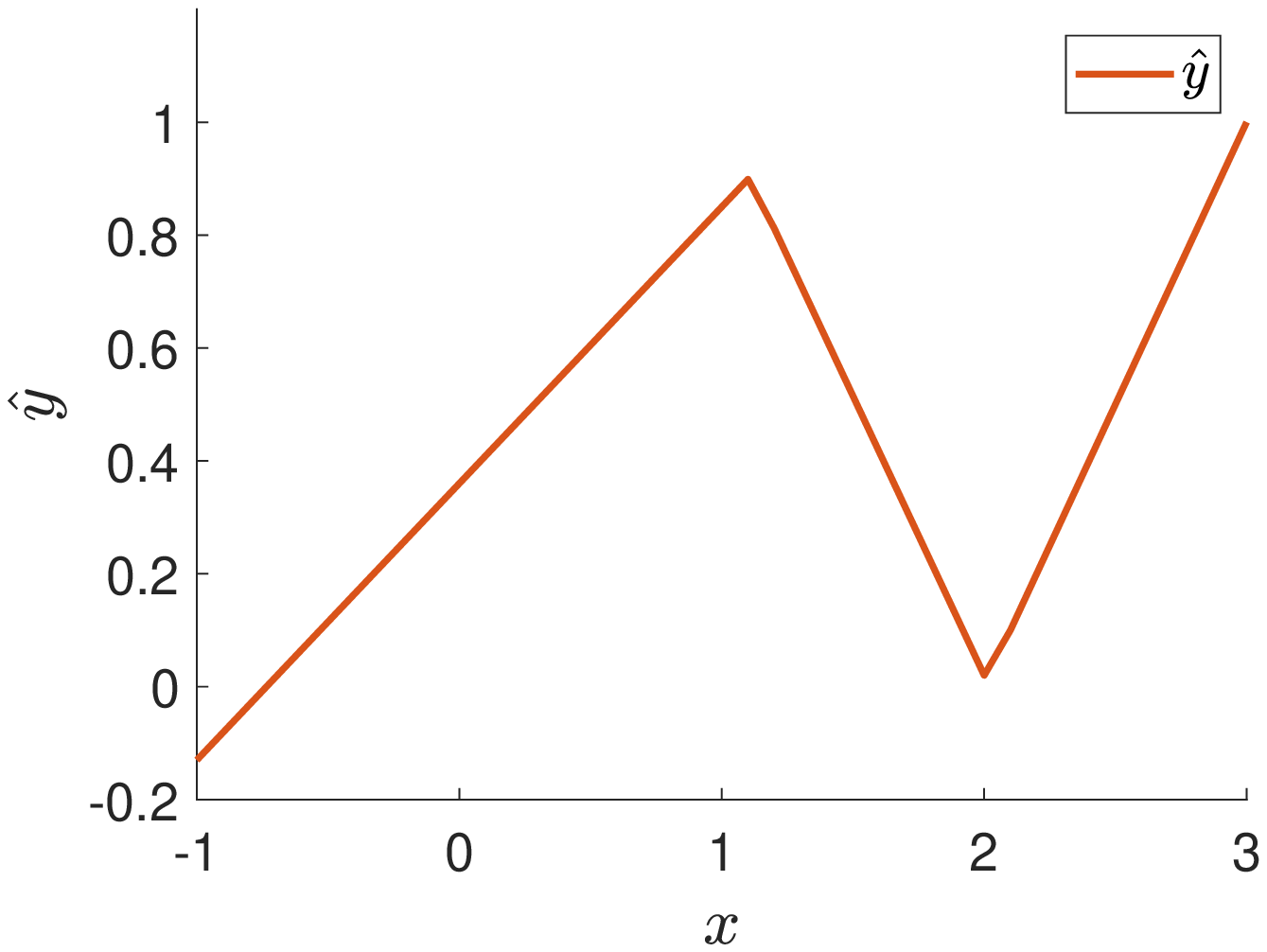} \label{subfig:d_1D}}
  \subfloat[$\hat{y}$ with with one concave and one convex neuron]{\includegraphics[width=0.35\textwidth]{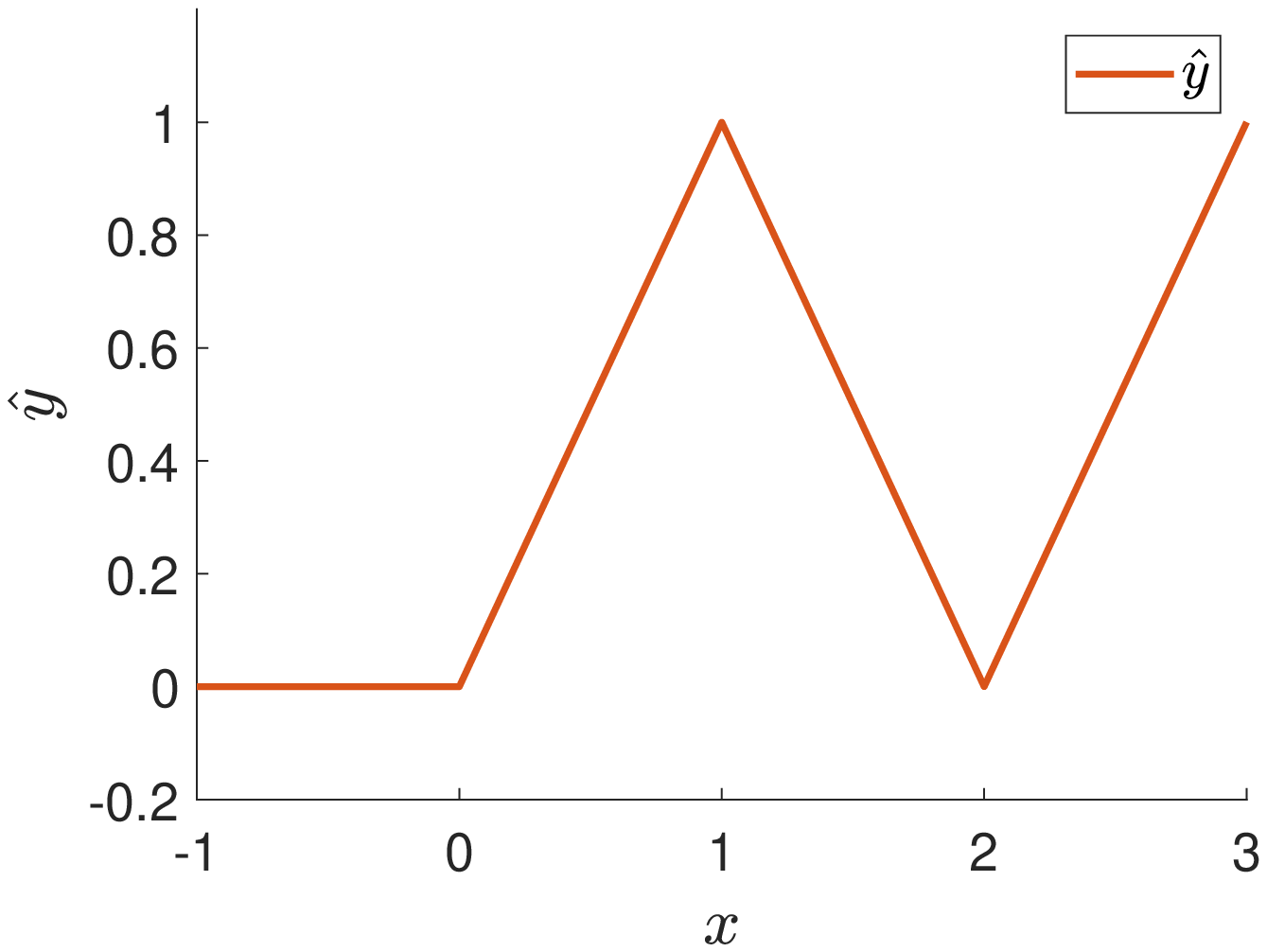} \label{subfig:e_1D}} \par
  \vspace{-10pt}
  \subfloat[Concave neuron $\hat{y}_{1 \min}$] {\includegraphics[width=0.35\textwidth]{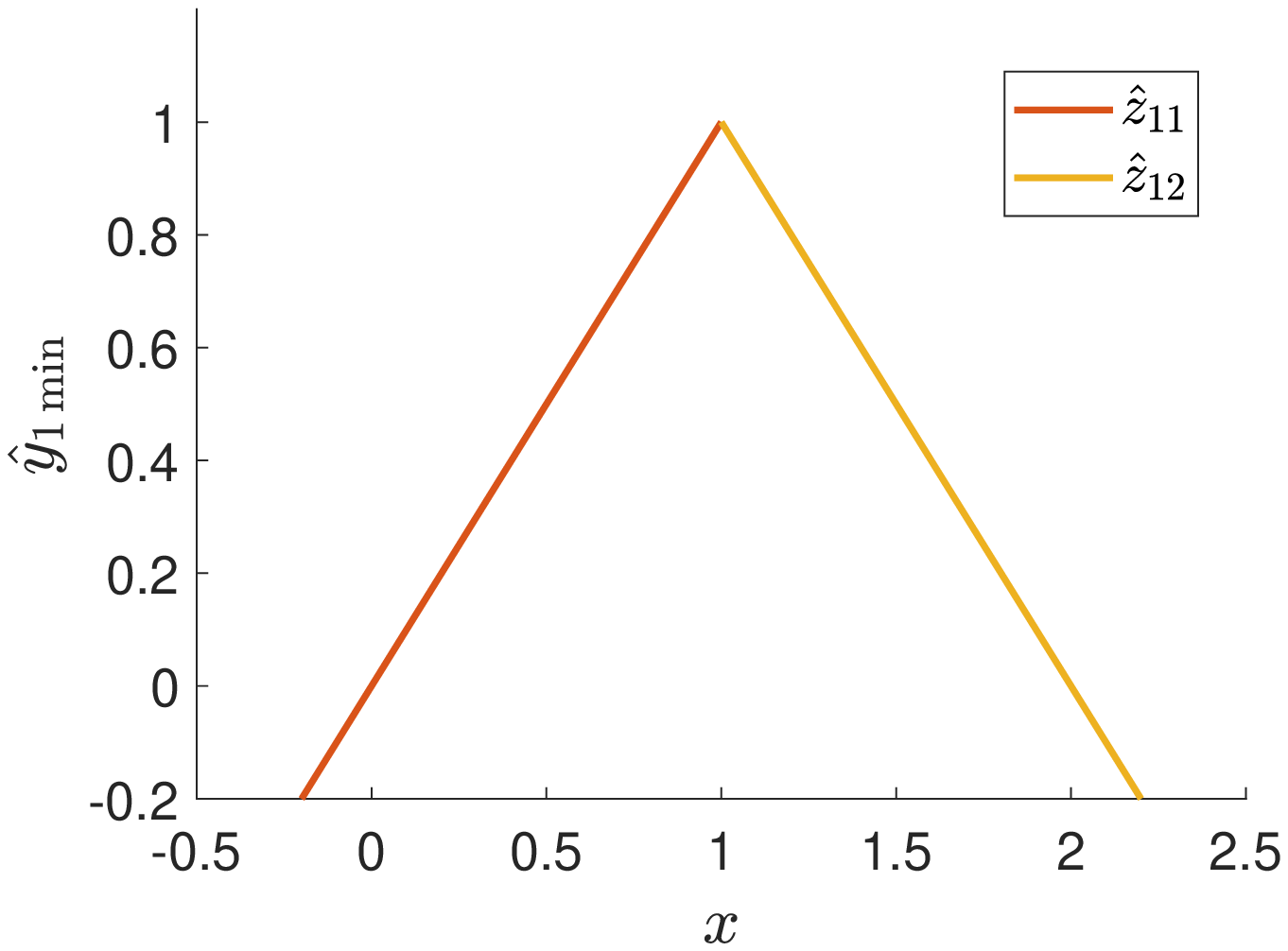} \label{subfig:f_1D}}
  \subfloat[Convex neuron $\hat{y}_{2 \max}$] {\includegraphics[width=0.35\textwidth]{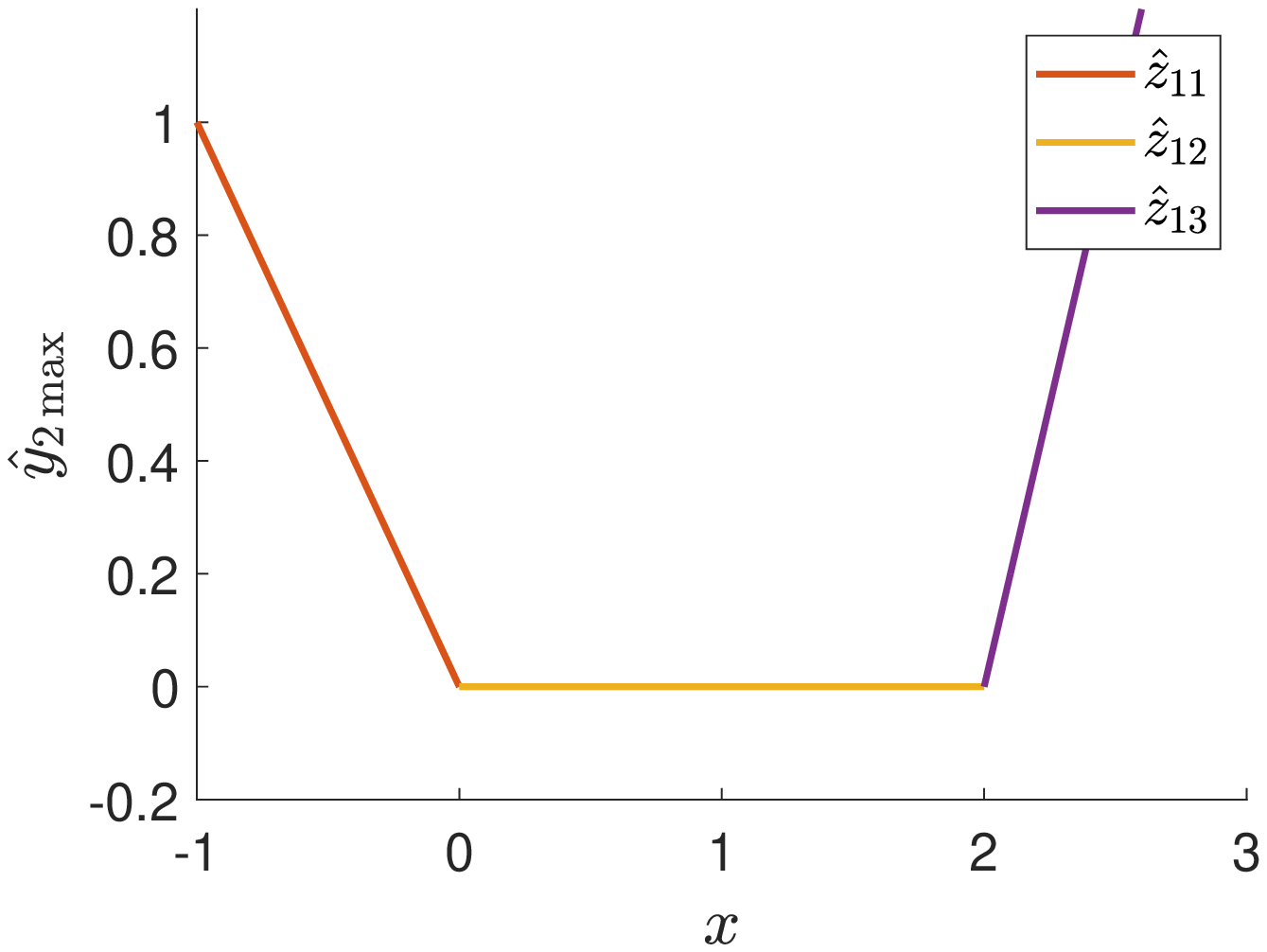} \label{subfig:g_1D}}
  \caption{Approximation of 1-dimensional target function by MinMax network}
\label{fig:ply}
\end{figure*}
}{1dex}




\Example{}{Figure \ref{fig:2D_example} shows how a MinMax network of Theorem \ref{th:1Dv2dis} evolves. We approximate the target function $y$ (\ref{eq:yx}) in subfigure \ref{subfig:a_2D} with the MinMax learning (\ref{eq:learning1Dv2}):

\begin{itemize}
    \item We start with one linear neuron in subfigure \ref{subfig:b_2D}. 
    \item The network then continuously inserts new basic neurons. Subfigure \ref{subfig:c_2D} depicts the approximated function with one neuron consisting of 3 linear basic neurons (\ref{eq:neuronj}). 
    \item Figure \ref{subfig:d_2D} shows the final result, which perfectly matches the target function. The final approximation $\hat{y}$ is the sum of the concave neuron $\hat{y}_{1 \max}$ and convex neuron $\hat{y}_{2 \max}$ (\ref{eq:neuronj}) shown in subfigure \ref{subfig:e_2D} and \ref{subfig:f_2D} respectively.
\end{itemize}

\begin{figure*}
  \subfloat[Target function $y$]{\includegraphics[width=0.45\textwidth]{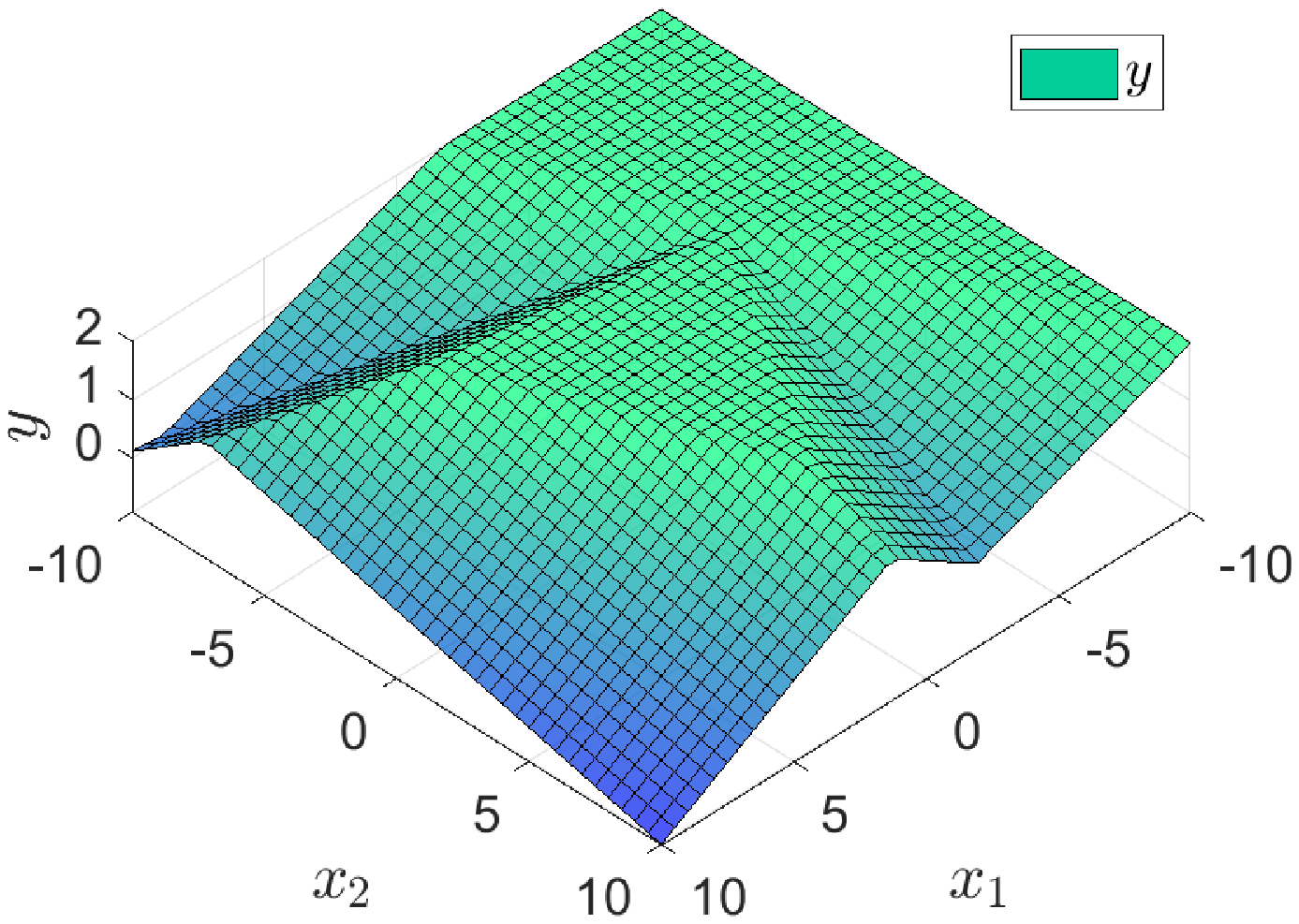} \label{subfig:a_2D}}
  \subfloat[$\hat{y}$ after initialization with one linear neuron]{\includegraphics[width=0.45\textwidth]{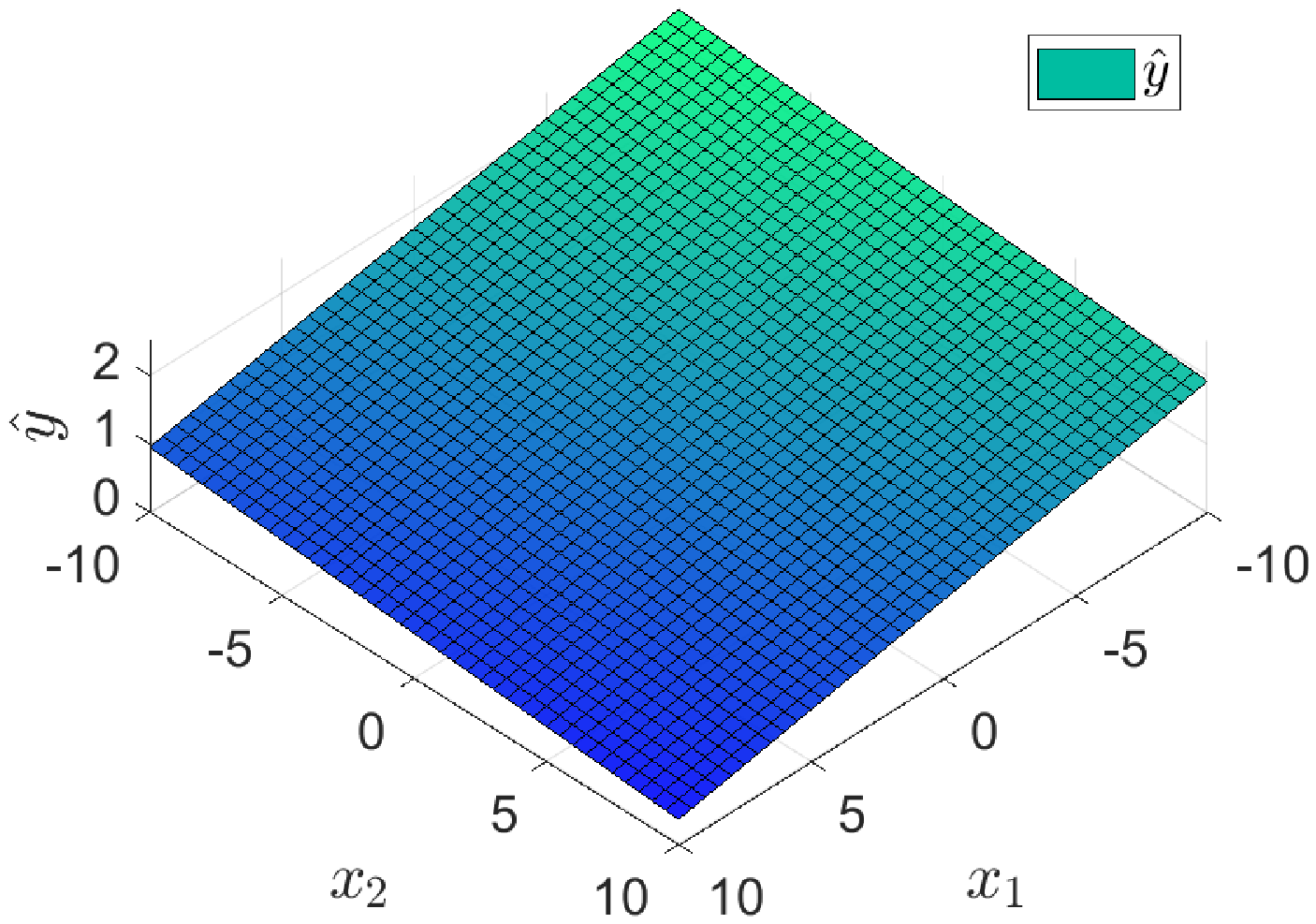} \label{subfig:b_2D}}\par
  \subfloat[$\hat{y}$ after two insertions of two basic neurons]{\includegraphics[width=0.45\textwidth]{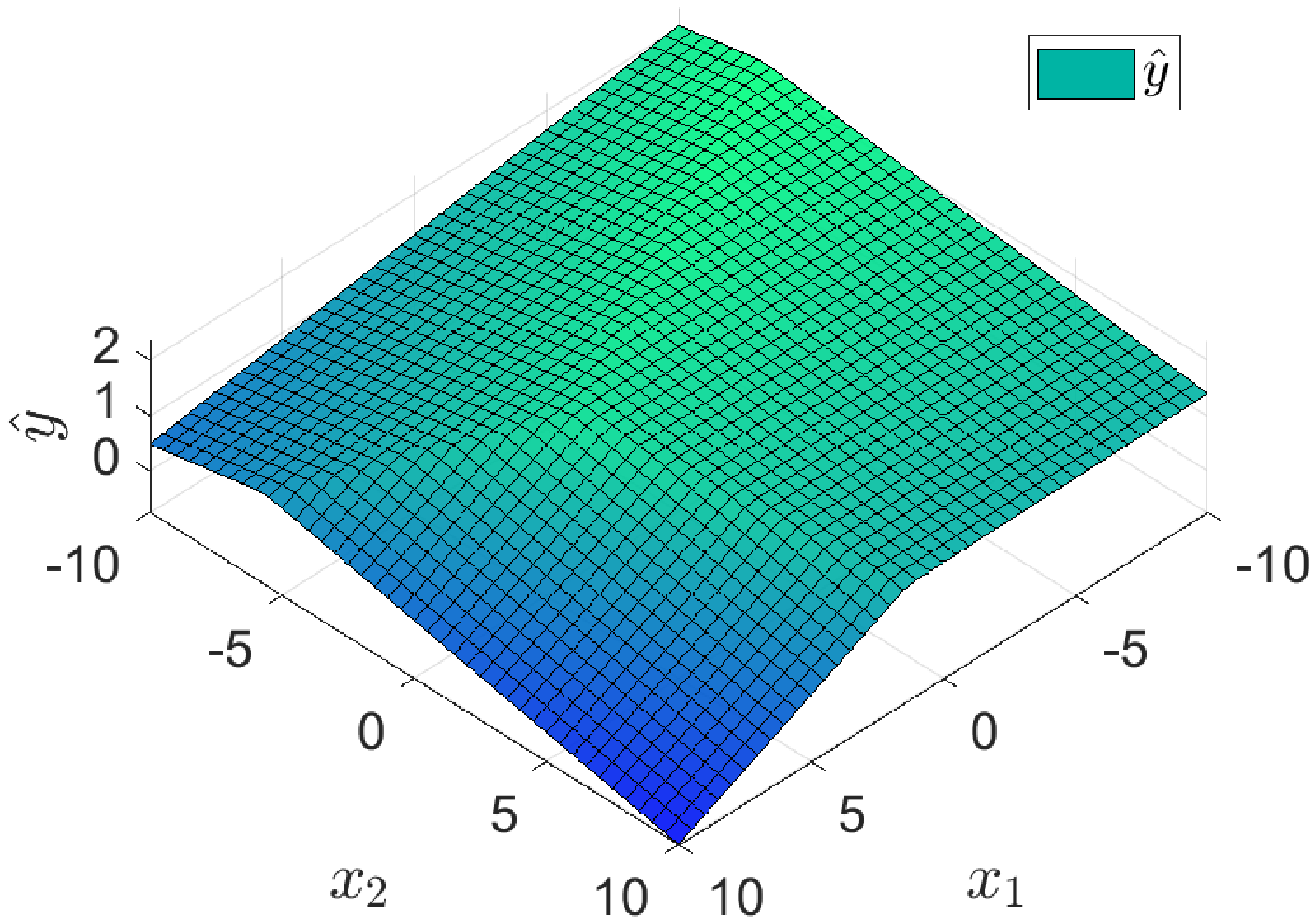} \label{subfig:c_2D}}
  \subfloat[$\hat{y}$ after full training with two neurons]{\includegraphics[width=0.45\textwidth]{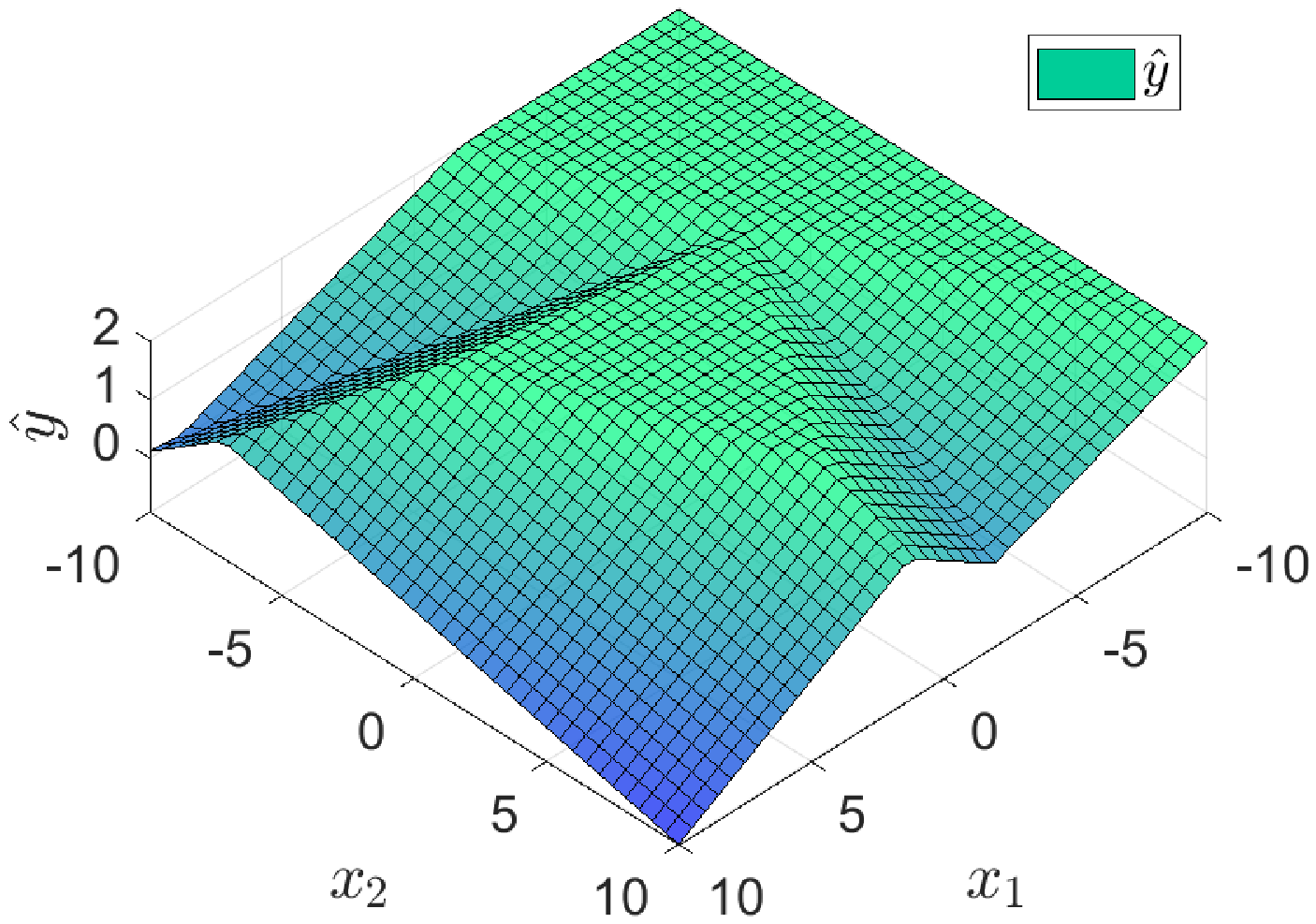} \label{subfig:d_2D}} \par
  \subfloat[Concave neuron $\hat{y}_{1 \max}$ of MinMax]{\includegraphics[width=0.45\textwidth]{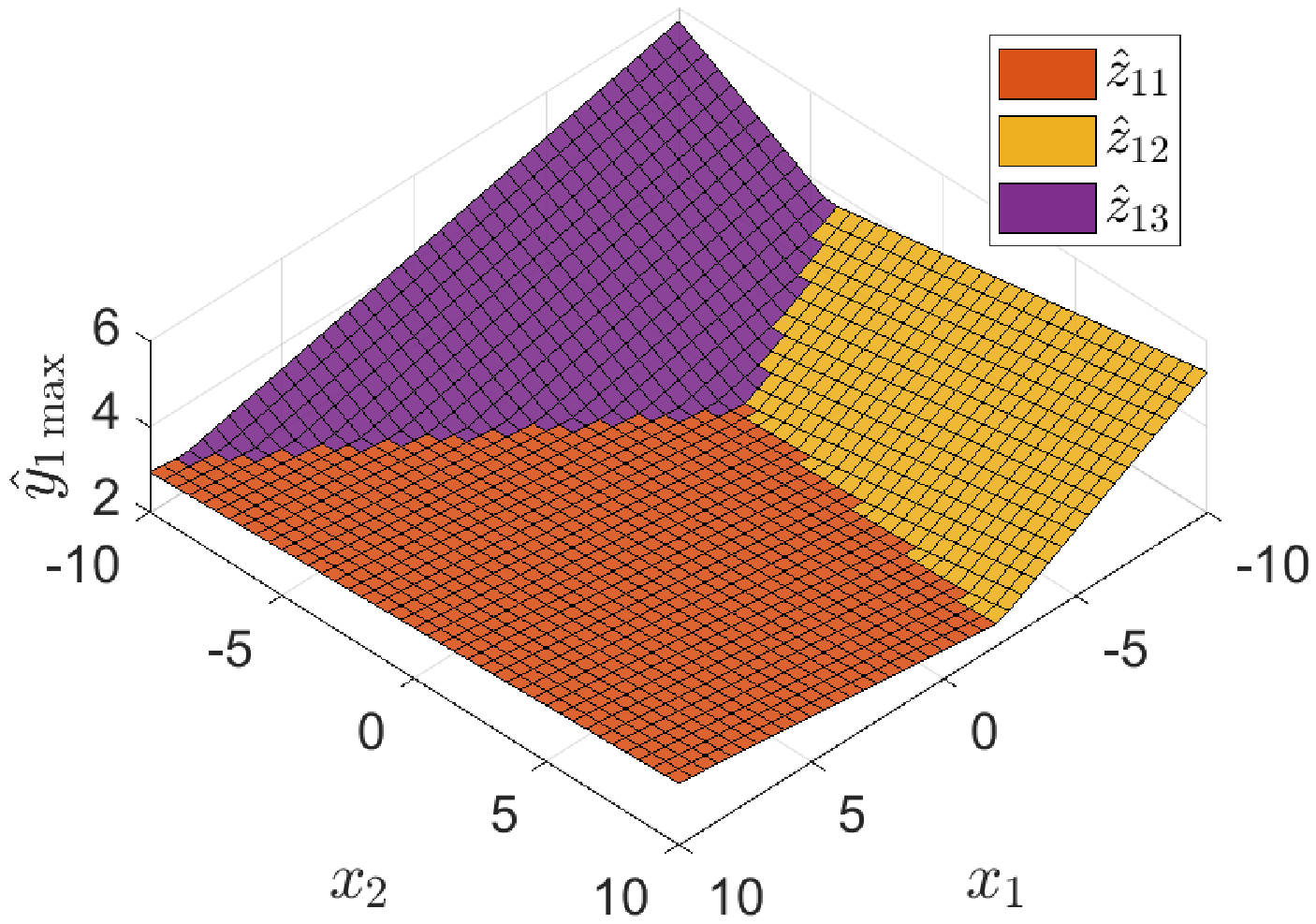} \label{subfig:e_2D}}
  \subfloat[Convex neuron $\hat{y}_{2 \min}$ of MinMax]{\includegraphics[width=0.45\textwidth]{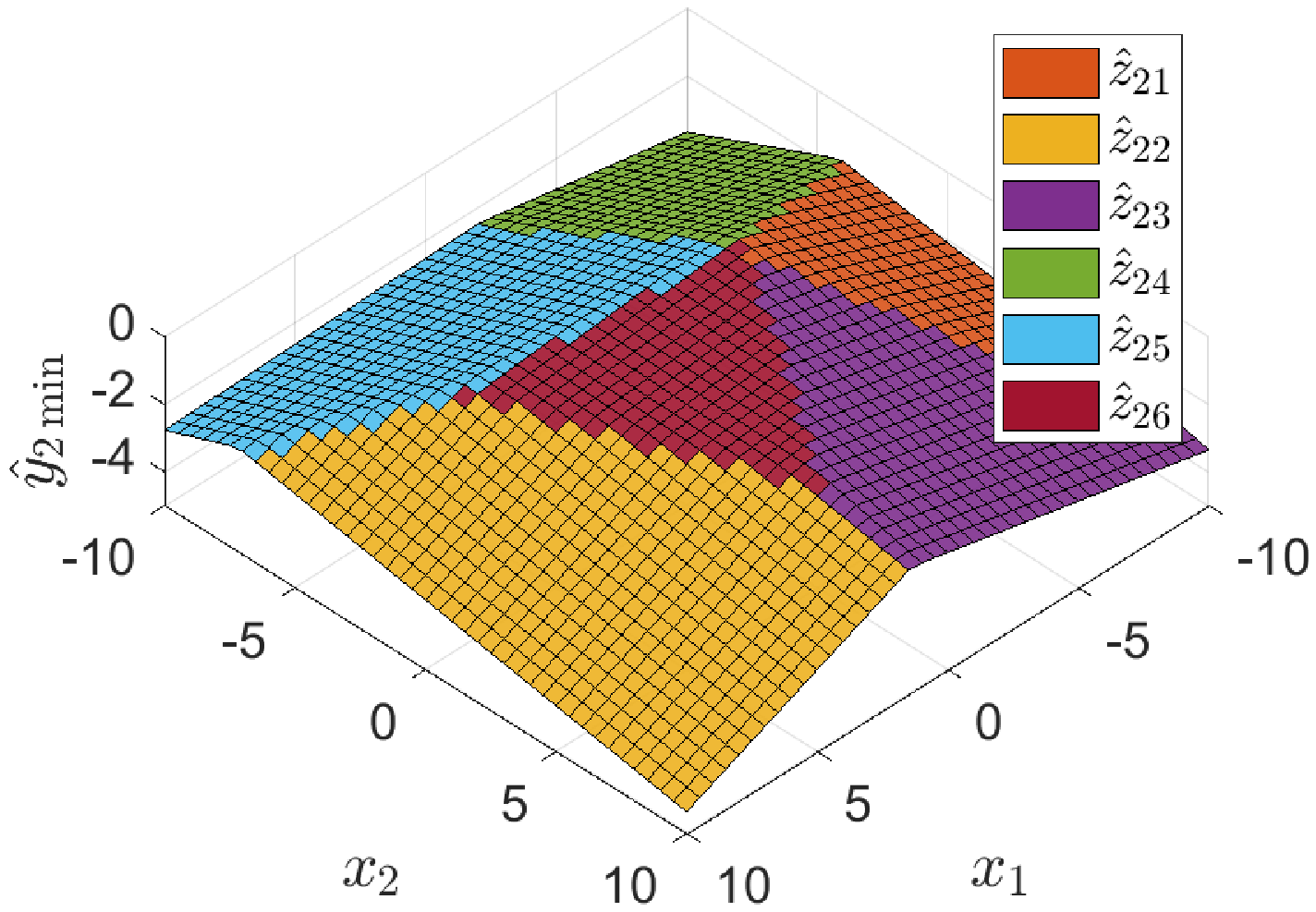} \label{subfig:f_2D}}
  \caption{Approximation of 2-dimensional target function by MinMax network} \label{fig:2D_example}
\end{figure*}
}{fff}

\Example{}{Figure \ref{fig:corner} shows the convergence of a MinMax Network of Theorem \ref{th:1Dv2dis} to a $8$-dimensional function 
\begin{equation}
    y({\bf x}) = \max_{n = 1, ..., 8}(\lvert x_n \rvert) \nonumber
\end{equation}

The network is initialized with a single neuron with random parameters $w_n \in [-0.5, 0.5]$ and further neurons are created after 100 iterations until the total cost $V$ (\ref{eq:V}) converges to zero. Steep drops in the error value indicate such insertions. Neurons are pruned if they become inactive or too similar to other basic neurons, leading to a network with the minimal required number of 16 basic neurons in one $\max$ neuron for the 16 linear surfaces.

\begin{figure*}
\begin{center}
\includegraphics[scale=0.75]{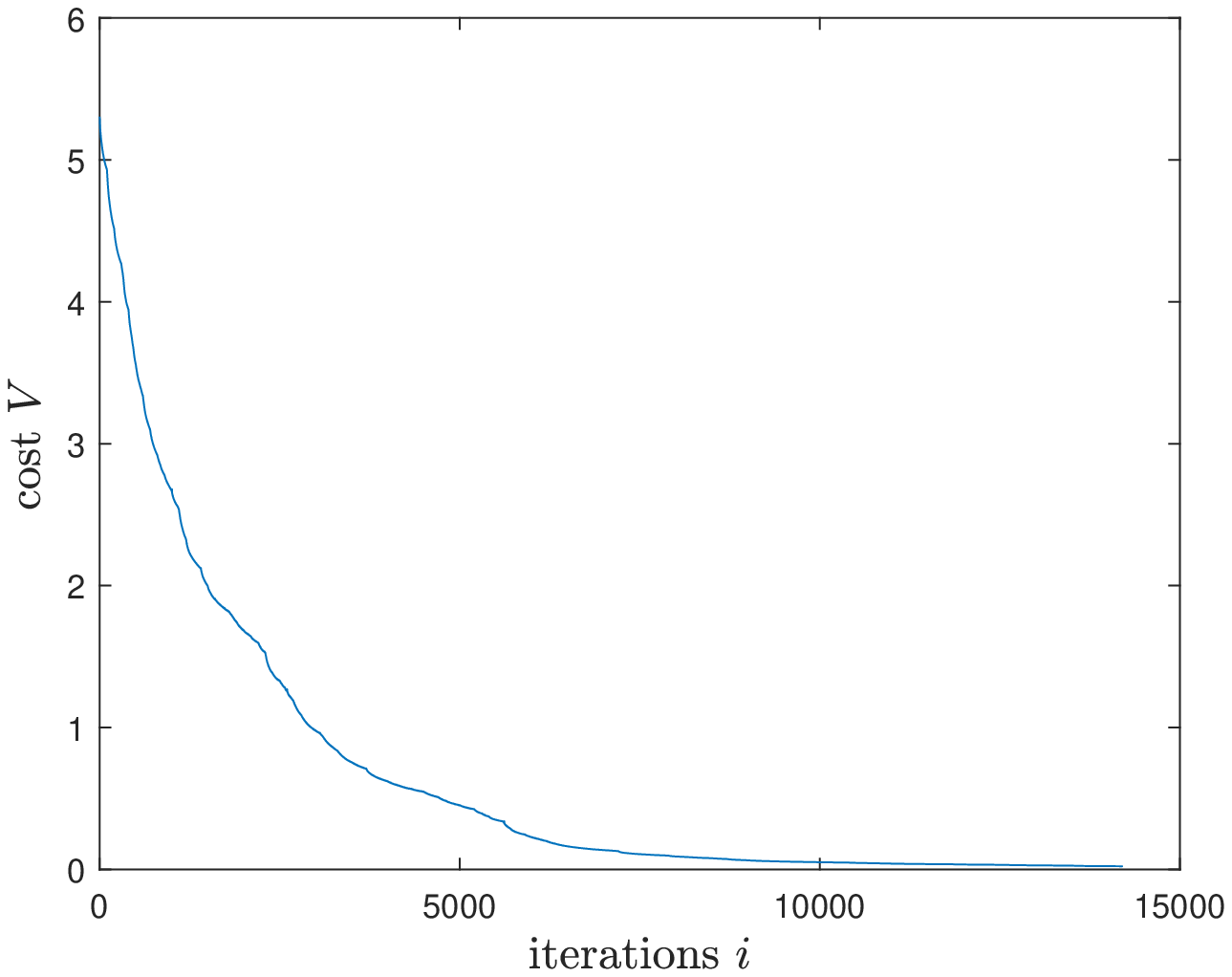}
\end{center}
\caption{Convergence of the cost $V$ over $i$ for the approximation of 16 surfaces in an 8-dimensional space by a MinMax network.}
\label{fig:corner}
\end{figure*}
}{}

\section{Summary}

This paper first extends discrete contraction theory to non-linear constrained systems in Theorem \ref{th:theoremFdis}. This allows to propose a new class of MinMax networks in Theorem \ref{th:1Dv2dis} for the learning of piece-wise linear functions: 
\begin{itemize}
    \item Possible instabilities or even non-unique solutions at the discontinuity between the linear regions are avoided by limiting ${\bf x}^{i+1}$ to its linear subspace with a Lagrangian constraint of Theorem \ref{th:theoremFdis}. This linear subspace may change at the next time instance since ${\bf x}^{i+1}$ then belongs to both neighbouring linear regions. 
    
    From a Contraction Theory perspective this constrained step parallelizes the virtual displacement $\delta {\bf x}^{i+1}$ to the edge, which is then orthogonal to the Dirac instability at the edge. Hence, it has no impact on the contraction rate under this constraint.
    \item  Saddles points or sub-optimal plateaus are avoided with a linear parametrization of the MinMax network. This is not possible with a deep network since the parametrization is highly polynomial.
\end{itemize}

As a result exponential convergence guarantees are given with Theorem \ref{th:1Dv2dis} for the discrete time learning of piece-wise linear functions.  

Although the learning was shown to be contracting in Theorem \ref{th:1Dv2dis} the remaining errors will not necessarily go to $0$ if a wrong topology was used. Hence the current research focus is to define finite neuron creation principles to find a correct topology of the MinMax network, where one error free solution exists.

Also, since each basic neuron is linear in Theorem \ref{th:1Dv2dis}, all linear estimation techniques can be exploited as e.g. computation of covariance matrices.

\end{document}